\documentclass{article}
\usepackage[final]{corl_2021}

\usepackage{amsmath}
\usepackage{amssymb}
\usepackage{booktabs}
\usepackage{graphicx}
\usepackage{inconsolata}
\usepackage{soul}
\usepackage{subcaption}
\usepackage[dvipsnames]{xcolor}

\usepackage{algorithm} 
\usepackage{algpseudocode} 

\newcommand{\MDP}{\mathcal M}           
\newcommand{\MState}{\mathcal S}        
\newcommand{\MUtterances}{\mathcal U}   
\newcommand{\MActions}{\mathcal A}      
\newcommand{\MTransition}{T}            
\newcommand{\MReward}{R}                

\def\Snospace~{\S{}}

\newcommand{\supp}{{Appendix}}

\hypersetup{
    colorlinks=true,
    linkcolor=orange,
    filecolor=magenta,      
    urlcolor=orange,
    citecolor=orange,
}

\title{LILA: Language-Informed Latent Actions}

\author{
  Siddharth Karamcheti\textsuperscript{\hypertarget{eqCont}{*}}, 
  Megha Srivastava\textsuperscript{\hypertarget{eqCont}{*}}, 
  Percy Liang, 
  Dorsa Sadigh \\
  Department of Computer Science, Stanford University\\
  \texttt{\{skaramcheti, megha, pliang, dorsa\}@cs.stanford.edu}
}
\begin{document}
\maketitle

\begin{abstract}
We introduce Language-Informed Latent Actions (LILA), a framework for learning natural language interfaces in the context of human-robot collaboration. LILA falls under the shared autonomy paradigm: in addition to providing discrete language inputs, humans are given a low-dimensional controller -- e.g., a 2 degree-of-freedom (DoF) joystick that can move left/right and up/down -- for operating the robot. LILA learns to use language to \textit{modulate} this controller, providing users with a language-informed control space: given an instruction like ``place the cereal bowl on the tray," LILA may learn a 2-DoF space where one dimension controls the distance from the robot's end-effector to the bowl, and the other dimension controls the robot's end-effector pose relative to the grasp point on the bowl. We evaluate LILA with real-world user studies, where users can provide a language instruction while operating a 7-DoF Franka Emika Panda Arm to complete a series of complex manipulation tasks. We show that LILA models are not only more sample efficient and performant than imitation learning and end-effector control baselines, but that they are also qualitatively preferred by users.\footnote{Additional visualizations and supplemental experiments can be found at the following webpage: \url{https://sites.google.com/view/lila-corl21}. Code can be found here: \url{https://github.com/siddk/lila}.}    
\end{abstract}


\section{Introduction}
\label{sec:introduction}
\vspace*{-0.03in}
Nearly a million American adults live with physical disabilities, requiring assistance for everyday tasks like taking a bite of food, or pouring a glass of milk \citep{taylor2018americans} -- assistance that robots could provide. Paradigms for efficient human-robot collaboration that strike a balance between robot autonomy and human control such as \textit{shared autonomy} \citep{dragan2013policy, argall2018autonomy, javdani2018shared, jeon2020sharedlatent} present a promising path towards building such assistive systems. Unlike full autonomy approaches that enforce a sharp separation between user intent and robot execution, falling prey to problems of sample efficiency and robustness, shared autonomy couples a human's input with automated robot assistance. Consider a kitchen or dining environment where a \textit{high-dimensional} (high-DoF) robot such as a wheelchair-mounted manipulator aids a human who may be physically unable to perform tasks requiring fine-grained manipulation. While the human can manually teleoperate the arm by fully controlling individual ``modes'', or separate degrees-of-freedom of the robot's end-effector, past work has shown this to be unintuitive, slow, and frustrating \citep{argall2018autonomy, herlant2016assistive}. Shared autonomy approaches such as \textit{learned latent actions} \citep{jeon2020sharedlatent, losey2020latent, losey2021latentactions} however, build intuitive low-dimensional controllers for high-DoF robots via dimensionality reduction.

Specifically, learned latent actions models learn \textit{state-conditioned} auto-encoders directly from datasets of (state, action) pairs; the encoder takes the current state and action, and compresses it to a latent action $z$ with the same dimensionality as the human control interface (e.g., 2-DoF). This is fed to a decoder to try to reconstruct the original high-dimensional action. While these approaches are reliable and sample efficient, they are limited by their reliance on just the current state: given tasks like ``grab the milk'' and ``shift the milk to the side'' that overlap in state space, these controllers fail as the models lack sufficient information to disambiguate behavior.

\begin{figure*}
    \centering
    \includegraphics[width=\textwidth]{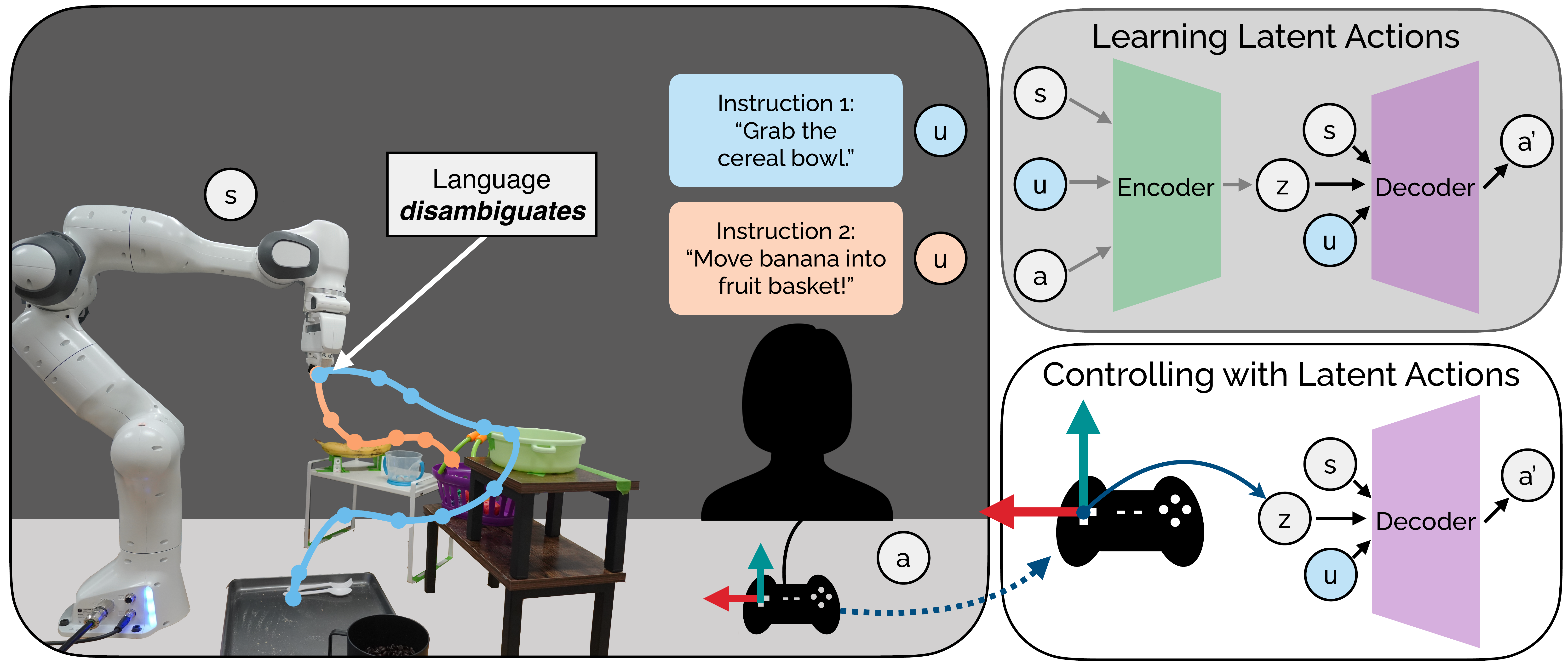}
    \vspace{0.03in}
    \caption{[\textbf{Left}] Our breakfast buffet environment with several diverse manipulation tasks. By providing both natural language input and low-dimensional joystick control [\textbf{Middle}], users \textit{disambiguate} between different tasks while retaining the ability to maneuver through the environment. This is enabled by our [\textbf{Right}] language-informed latent actions (LILA) models that use auto-encoders to learn language \& state-conditioned low-DoF latent spaces for meaningful control.}
    \label{fig:front}
\end{figure*}

To address this concern, we consider incorporating natural language within this framework to add an additional conditioning variable for structuring the control space. Prior work integrates language in robotics settings for similar purposes within the full autonomy paradigm \citep{tellex2011understanding, arumugam2017accurately, anderson2018butd, hermann2017grounded, stepputtis2020lcil}. Unfortunately, these approaches suffer from poor sample efficiency, failure recovery, and generalization; many issues that shared autonomy methods including learned latent actions seek to address. By joining language and latent actions, a user can express an utterance $u =$ \textit{``grab the cereal bowl''} and obtain a control space that is both state and language conditioned (\autoref{fig:front} [Right]). 

We introduce Language-Informed Latent Actions (LILA), a framework for incorporating language into learned latent actions. Key to LILA is the principle that language \textit{modulates} a user's low-level controller based on their provided utterance; as intuition, given the utterance \textit{``grab the cereal bowl''} as in \autoref{fig:front}, our assistive robot might learn a semantically meaningful, low-dimensional (2-DoF) control space where one dimension (one joystick axis) may control the distance from the robot's end-effector to the cereal bowl, whereas the other might control the angle of the end-effector relative to the bowl such that the robot's gripper can obtain a solid grasp of the object. Other utterances can modulate the controller in similar ways -- ``pour the milk into the cup'' might result in a learned control space where one joystick dimension controls the jug's pouring angle, while the other may control its height. Language not only serves as a natural means for a human to communicate their intent to the robot, but also helps \textit{disambiguate} across a wide variety of objectives as well, by inducing language and state-conditioned control spaces.

A core part of our method is its ability to handle diverse, realistic language. To this end, we collect a small, crowdsourced dataset of natural language descriptions to describe each of our training demonstrations; we use this real, natural language as the only input while training our models. To allow for out-of-the-box generalization to novel user utterances such as those that describe similar behaviors but with different words or phrases, we tap into the power of pretrained models \citep{liu2019roberta, reimers2019sentence}. We perform a \textit{comprehensive user study} across 10 users who use natural language and our learned LILA controllers to complete a variety of diverse manipulation tasks in a simplified assistive ``breakfast buffet'' setting. Our results show that LILA models are not only more reliable, performant, and sample efficient than fully autonomous imitation learning and fully human-driven end-effector control baselines, but are qualitatively preferred by users as well.

\vspace*{-0.07in}
\section{Related Work}
\label{sec:related-work}
\vspace*{-0.03in}
We build LILA within a shared autonomy framework \citep{dragan2013policy,gopinath2016human}, applied to assistive teleoperation \citep{ciocarlie2009hand, jain2019probabilistic}. We additionally build off of work at the intersection of language and robotics \citep{luketina2019survey, matuszek2018groundedlang, tellex2020robonlp}.

\paragraph{Shared Autonomy \& Assistive Teleoperation.} Shared autonomy casts robot control as a collaborative process between humans and robots \citep{dragan2013policy, javdani2018shared, gopinath2016human, newman2018harmonic}. While other work focuses on ``blending'' human inputs with possibly task-agnostic policies within the same action space \citep{reddy2018shared, schaff2020residual}, in this work, we focus on assistive teleoperation, where a user is provided a low-dimensional controller (e.g., a joystick, sip-and-puff device) to directly control a high-dimensional robot manipulator. Using these controllers for end-effector control -- e.g., via operational space control \citep{khatib1987osc} -- is incredibly difficult, requiring frequent mode-switching to control specific robot DoFs \citep{argall2018autonomy,wilson1996relative}. Instead, we adopt \textit{learned latent actions} \citep{jeon2020sharedlatent, losey2020latent, losey2021latentactions, li2020intuitive, karamcheti2021vla} a framework that uses conditional auto-encoders \citep{doersch2016tutorial} to learn task-specific latent ``action'' spaces from demonstrations. These latent spaces match the dimensionality of the low-DoF interface and provide semantically meaningful control. However, existing methods fail to differentiate between tasks with overlapping states, hindering the ability to perform diverse behaviors in a workspace (e.g., manipulating a jug of milk in different ways -- pouring, placing in the fridge, etc.). In this work, we use language for \textit{disambiguation}; users naturally speak their intent, conditioning latent action models to produce intuitive control spaces that align with user objectives.

\paragraph{Language-Informed Robotics.} A variety of methods have sought to combine language and robotics, spanning approaches that map language to planning primitives \citep{tellex2011understanding,kollar10directions,matuszek2012learning,artzi2013weakly,karamcheti2017draggns}, perform imitation learning from demonstrations and instructions \citep{anderson2018butd, blukis2018following, wang2019rcm, lynch2020grounding}, and pair language instructions with reward functions for reinforcement learning \citep{hermann2017grounded, chaplot2018gated, hill2020human}. Other approaches use language in more nuanced ways, such as learning language-conditional reward functions directly \citep{macglashan2015grounding, fu2019lang2goals, bahdanau2019reward}, or within adaptive frameworks, where language is used to correct or define new behavior \citep{coreyes2019guiding, karamcheti2020decomposition}. This list is not exhaustive; we present further discussion -- including approaches that combine language with other modalities -- in the \supp{}. However, all these approaches fall within full autonomy: after providing an instruction, human users cede control over to the robot policy, which then takes the actions necessary to perform a task. 

While robots trained with these approaches can perform diverse tasks and generalize to new instructions, it is not without cost. Paramount is sample efficiency; imitation learning approaches often require hundreds to thousands of demonstrations for learning to navigate \citep{wang2019rcm,anderson2018vision}, and reinforcement learning approaches can require millions of episodes of experience to learn robust policies \citep{hermann2017grounded, chevalierboisvert2019babyai}. Whereas coarse behaviors are easy to learn, learning to recover from slight deviations from the training data, or to perform precise motions in a sequence, is incredibly difficult. By casting our approach, LILA, within a shared autonomy framework instead, we intelligently offload these parts that are harder for robots -- but easier and intuitive for humans -- onto the user.

\begin{figure*}[t]
    \centering
    \includegraphics[width=\textwidth]{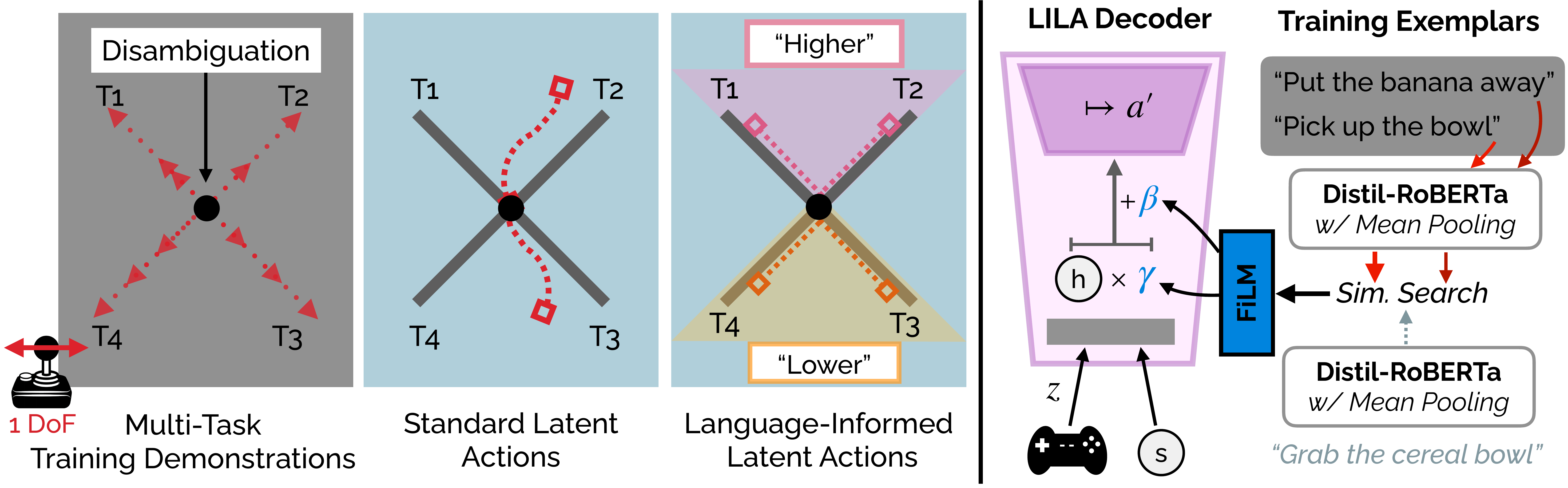}
    \vspace{0.03in}
    \caption{[\textbf{Left}] Stylized example: navigating toward four points on a cross with a 1-DoF latent action, where disambiguation is required. Standard latent action models fail, while LILA accurately reaches the corners with the help of language. [\textbf{Right}] LILA decoder architecture. We embed an utterance using a pretrained language model, then identify the closest exemplar in the training set via similarity search. We feed the embedding for this exemplar through a feature-wise linear modulation, or FiLM \citep{perez2018film}, layer that fuses language and state representations within the decoder.}
    \label{fig:xarch}
\end{figure*}

\vspace*{-0.07in}
\section{Formalizing Language for Assistive Teleoperation}
\label{sec:framing}
\vspace*{-0.03in}
\paragraph{Formalism.} We formulate a user's objective, or task, (on a \textit{per-user basis}) as a fully-observable, language-augmented, Markov Decision Process (MDP) $\MDP$ defined by the tuple $(\MState, \MUtterances, \MActions, \MTransition, \MReward, \gamma)$ similar to prior work in language-conditional robotics \citep{arumugam2017accurately, blukis2018following, coreyes2019guiding}. Let $u \in \MUtterances$ denote a user's language utterance provided at the start of each episode, where $\MUtterances$ is the full set of language utterances a user could provide to a robot. Let $s \in \MState \subseteq \mathbb{R}^n$ be the robot's state, and $a \in \MActions \subset \mathbb{R}^m$ be the robot's action: taking action $a$ in state $s$ results in a next state $s'$ according to the transition function $\MTransition(s, a)$. Given the language utterance $u$, the user implicitly defines a reward function $\MReward(s, u, a) \in \mathbb{R}$; the human and robot collaboratively maximize this reward subject to discount factor $\gamma \in [0, 1]$.

\paragraph{Problem Statement.} This MDP forms the basis of a shared autonomy task wherein a human is equipped with a low-dimensional control interface for the robot. Let $z \in \mathcal{Z} \subset \mathbb{R}^d$ where $d \ll m$ be the human's control input to the robot, such as the $d=1$ DoF controller in \autoref{fig:xarch}. Previous work on learning latent actions for assistive teleoperation \citep{jeon2020sharedlatent, karamcheti2021vla} learn a decoder $\text{Dec}(s, z) : \MState \times \mathcal{Z} \rightarrow \MActions$ that maps user low-dimensional inputs $z \in \mathcal{Z}$ and current state $s \in \MState$ to a high-dimensional action $a \in \MActions$. However, in situations where state-conditioning is not enough to disambiguate a users' intent, too low of a control input dimension $d$ may lead to failure. Recalling the milk jug example, we have multiple different behaviors we could execute if the end-effector were next to the milk. For example, one might want to pick up the jug, shift it to the side, pour it, etc; conditioning only on the state with a 2-DoF action space is not enough to recover all possible behaviors.

Instead, we aim to learn $\text{Dec}(s, u, z) : \MState \times \MUtterances \times \mathcal{Z} \rightarrow \MActions$ that takes the user's control input \textit{and} utterance $u$, and predicts the high-DoF action that matches the user's objective. The utterance $u$ acts as additional conditioning information, producing control spaces that depend on both language and state; this circumvents the disambiguation problem above.

\vspace*{-0.07in}
\section{Language-Informed Latent Actions (LILA)}
\label{sec:lila}
\vspace*{-0.03in}
We are given a dataset of demonstrations, where each demonstration contains an utterance $u$ and a trajectory $\tau = \{s_0, a_0, s_1, a_1 \ldots s_T\}$. We split each demonstration into triples of $(u, s_i, a_i)$ and use these to learn a conditional auto-encoder, consisting of a language-conditional encoder $\text{Enc}$: $\MState \times \MUtterances \times \MActions  \rightarrow \mathcal{Z}$ that maps to a latent $z$ and a decoder $\text{Dec}$: $\MState \times \MUtterances \times \mathcal{Z} \rightarrow \MActions$ that attempts to reconstruct the original action $a$. We minimize the mean-squared error between the predicted and the original action:
\begin{equation}
L_{\text{Enc}, \text{Dec}} = \frac{1}{N}\sum_{i=1}^N(\text{Dec}(s_i, u_i, \text{Enc}(s_i, u_i, a_i)) - a_i)^2
\end{equation}
We next discuss how we integrate language into the architecture of the encoder and decoder.

\begin{table}[t]
\centering
    \renewcommand\tabcolsep{3pt}
    \resizebox{\textwidth}{!}{
    \begin{tabular}{lcp{5cm}p{5cm}}
    \toprule
       \textbf{Task Name}& \multicolumn{1}{c}{\textbf{Success}} & \multicolumn{1}{c}{\textbf{Example User Study Input}} & \multicolumn{1}{c}{\textbf{Mapped Training Data}} \\
    \toprule
    \textbf{\texttt{Pick Banana}} & 100\% & \emph{yellow in purple} & $\rightarrow$ \texttt{pick up the yellow banana and place it into the purple basket} \\
    \midrule
    \textbf{\texttt{Pick Fruit Basket}} & 100\% &  \emph{bring basket to center of pan} & $\rightarrow$ \texttt{place the basket onto the tray} \\
        \midrule
    \textbf{\texttt{Pick Cereal}} & 100\% &\emph{go to the left side of the cream bowl, go down, grab the cereal bowl, and place it on the try}  & $\rightarrow$ \texttt{grab the cereal bowl and put it on the tray} \\    
    \midrule
    \textbf{\texttt{Pour Bowl}} & 67\% &  \emph{pick up the cup of marbles and pour them into the cereal bowl} & $\rightarrow$ \texttt{pick and pour the cup of white balls into the bowl of cereal} \\    
    \midrule
    \textbf{\texttt{Pour Cup}} & 100\% &  \emph{pick up the clear cup with marbles in it and pour it in the black mug with the coffee beans in it} & $\rightarrow$ \texttt{pick up the cup and pour the contents in the mug} \\
    \bottomrule
    \end{tabular}
    }
    \medskip
    \captionof{table}{Example utterances provided by study participants paired with the retrieved exemplar per \autoref{subsection:integrating-language}. Success rate refers to the percentage of the time a user study utterance (over all utterances in the study) was grounded to the correct task via our retrieval method. As each participant only attempted 2 tasks, success rate can fluctuate significantly, as is the case with the \textbf{\texttt{Pour Bowl}} task.}
    \label{tab:table}
\end{table}

\subsection{Integrating Language within the Latent Actions Architecture}
\label{subsection:integrating-language}

We implement the encoder and decoder as multi-layer feed-forward networks, with the ReLU activation as in prior work \citep{losey2020latent,karamcheti2021vla}. We focus on the decoder here, but the encoder is symmetric. For the decoder, we first concatenate the robot state $s$ and latent action $z$, then feed the corresponding vector through multiple ReLU layers (usually 2-3), upsampling to produce the high-dimensional robot action. We next discuss how to incorporate language within this simple scaffold.

Pretrained language models such as BERT, T5, and GPT-3 \citep{devlin2019bert,raffel2019exploring,brown2020gpt3} have revolutionized NLP, providing powerful language representations. Inspired by their success when applied to robotics and reinforcement learning tasks \citep{hill2020human, marzoev2020unnatural, ku2020rxr}, we use a distilled RoBERTa-Base model \citep{liu2019roberta}, from \texttt{Sentence-Transformers} \citep{reimers2019sentence} to encode utterances. This model is fine-tuned on a corpus of paraphrases, allowing it to pick up on sentence-level semantics. We generate utterance embeddings by performing mean-pooling over token embeddings for an utterance, as in prior work~\citep{hill2020human,marzoev2020unnatural}. We incorporate these embeddings using feature-wise linear modulation (FiLM) layers \citep{perez2018film} that fuse language information with other features $h$ by mapping language embeddings to parameters $(\gamma, \beta)$ of an affine transformation: $h' = \gamma * h + \beta$ (\autoref{fig:xarch}). Notably, this $h$ is the representation received after feeding the state and latent action $(s, z)$ through the \textit{first} layer of the decoder as described above. Once the language transforms $h \rightarrow h'$, we feed $h'$ to the subsequent layers of the decoder.\footnote{Implementation can be found in the open-source code repository: \url{https://github.com/siddk/lila}.}

\paragraph{Nearest-Neighbor Retrieval at Inference.} A major concern for work in language-conditioned robotics is generalizing to novel language inputs. While it may be unreasonable to expect generalization to completely new tasks, for user-facing systems with a clear set of behaviors seen at train time as in our work, there is an expectation that any language-informed system is capable of handling moderate variations of utterances from the training set. To do this, adding linguistically diverse data has been the gold standard \citep{anderson2018vision, ku2020rxr}; however, a new class of approaches have emerged that sidestep additional data requirements by tapping into the potential of pretrained language models \citep{karamcheti2020decomposition,marzoev2020unnatural}. These approaches frame language interpretation at inference, when interfacing with real users, as a \textit{retrieval} problem: each new user utterance $u'$ is embedded(with the same pretrained model as above, then used to query a nearest neighbors store containing all training exemplars; once the nearest neighbor $u_i$ has been identified, it replaces $u'$ as an input to LILA.

The key benefit of such an approach is the minimal mismatch between train and test language inputs: all ``test inputs'' are drawn from the training set. This does mean, however, that user utterances that describe new tasks, or are otherwise unachievable also get mapped to language seen at training. While this limits the ability to perform novel tasks, it again highlights the benefits of the shared autonomy paradigm -- doubly so, considering the cost of a mistake in an assistive domain like the one we consider in this work: if, while providing control inputs to the robot, a user feels the robot is not acting in alignment with the user's desired objective, they can always stop execution.

\vspace*{-0.07in}
\section{User Study}
\label{sec:user-study}
\vspace*{-0.03in}
We evaluate LILA with a real-world user study on a 7-DoF Franka Emika Panda Arm, on a series of 5 complex manipulation tasks. Each user is provided a 2-DoF joystick for control. We compare against a non-learning, end-effector (EE) control baseline where users ``mode switch,'' controlling the velocity of 2 axes of the end-effector pose at a given time -- [(X, Y), (Z, Roll), (Pitch, Yaw)]. Language utterances $u$ are typed into a text console for simplicity; future work could extend this work by using off-the-shelf speech recognition systems. We also compare against a fully autonomous imitation learning (IL) strategy where users solely provide language inputs, and the robot attempts to perform a task without additional input. Ostensibly missing is a \textit{no-language} variant of the latent actions model, in keeping with prior work; however, upon evaluating this model, we found it to be unintuitive and unable to make progress or solve any task, so we omit it from our user study. However, further experiments and analysis can be found in the \supp{}.\footnote{Experiments with the no-language baseline and extra analysis showing the necessity of extra conditioning information can be found here: \url{https://sites.google.com/view/lila-corl21/home/no-lang-baseline}}

\paragraph{Environment.} \autoref{fig:front} shows our ``breakfast buffet'' setting, a scaled down version of an assistive feeding domain. We define 5 tasks: 1) \textbf{\texttt{Place Banana}}: placing the banana in the purple fruit basket, 2) \textbf{\texttt{Place Basket}}: grasping the purple basket by the handles and dropping it on the tray, 3) \textbf{\texttt{Place Bowl}}: grasping the green cereal bowl by its edge and moving it to the tray, 4) \textbf{\texttt{Pour Bowl}}: pouring the blue cup of marbles (a proxy for milk) into the cereal bowl positioned on the tray, and 5) \textbf{\texttt{Pour Cup}}: pouring the blue cup of marbles into the yellow coffee cup. \autoref{fig:front} shows idealized example trajectories for the \textbf{\texttt{Place Bowl}} (blue) and \textbf{\texttt{Place Banana}} (orange) tasks. 

\begin{figure*}
    \centering
    \includegraphics[width=\textwidth]{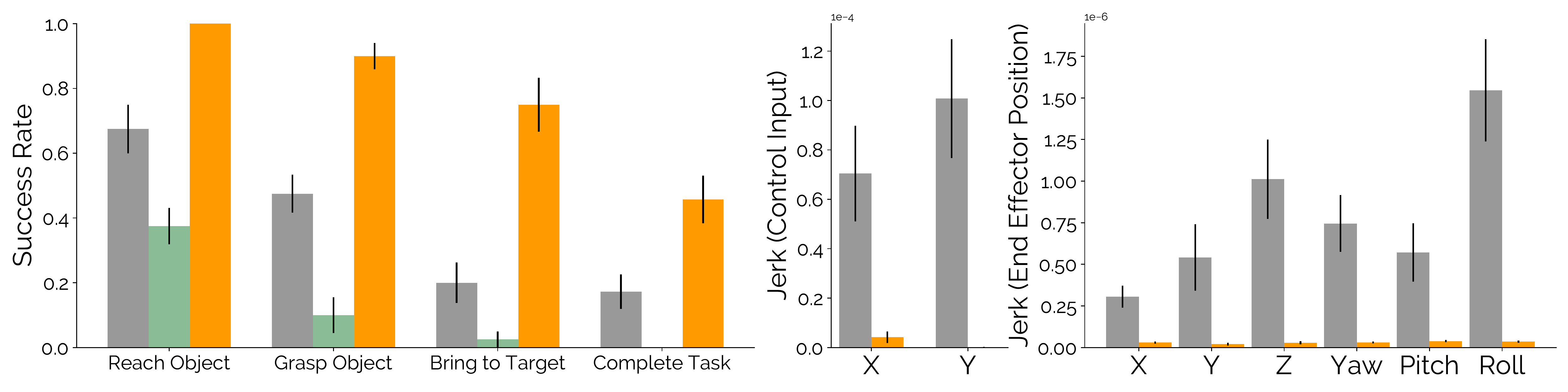}
    \vspace{0.03in}
    \caption{Quantitative Results. We average success rate [\textbf{Left}] across all sub-tasks for each control method, and find that LILA is significantly ($p < 0.05$) more performant. However, the steep drop in performance when completing the full task shows the difficulty of fine-grained control. We also calculate jerk as an indicator of controller smoothness, for both user control inputs [\textbf{Middle}] and end-effector position [\textbf{Right}]. Averaged across tasks and users, we find LILA leads to significantly smoother control for users than end-effector control.}
    \label{fig:quantitative}
\end{figure*}

These tasks vary in difficulty, requiring precise grasping and dexterous manipulation. We evaluate partial success based on how many of the following 4 subtasks users are able to complete: 1) \textbf{\texttt{Reaching}}: touching the desired object, 2) \textbf{\texttt{Grasping}}: executing a successful grasp, 3) \textbf{\texttt{Bring to Target}}: successfully transporting the manipulated object, and 4) \textbf{\texttt{Task Completion}}.

\paragraph{Demonstration Collection.} Both LILA and IL models require learning from (language, demonstration) pairs for all 5 tasks. We collect demonstrations kinesthetically as in prior work \citep{losey2020latent, karamcheti2021vla}, recording joint states at a fixed frequency. We initially collected 15 demonstrations per task for each method. However, on testing the IL model, we found it incapable of performing even rudimentary reaching behaviors. To give IL the best chance, we collected twice the number of demonstrations (30 per task; 150 total), requiring an extra 2 hours of labor.

\paragraph{Crowdsourcing Language Annotation.} To build a \textit{natural} language interface for human-robot collaboration, we collect language annotations for each task by crowdsourcing utterances. Our goal was to capture the diverse ways users may refer to the objects and actions our tasks entail without any additional information, simulating a real user interacting with our environment for the first time. We recruited 30 workers on Amazon Mechanical Turk, showing only a video of a recorded demonstration, and asked them to provide \textit{``a short instruction that you would want to provide the robot to complete this task independently in the box below.''}. However, this procedure resulted in some annotations containing ``spam'', or extremely out-of-domain text. To address this without introducing our own bias on what constitutes ``spam'', we filtered the data to identify workers who consistently provided ``noisy'' annotations, measured by the cosine distance between the sentence embedding (using any pretrained embeddings) of an annotator's provided text and the \textit{average} sentence embedding aggregated over all other annotators for a given video. We used annotations from the 15 least ``noisy'' annotators under this metric as our ground-truth utterances. Further details, as well as example ``spam'' annotations that were filtered out, are in the \supp{}. \autoref{tab:table} provides examples of crowdworker utterances from our final dataset (rightmost column).

\paragraph{Participants \& Procedure.} We conducted our study with a participant pool of 10 university students (5 female/5 male, age range $23.2 \pm 1.87$). Four subjects had prior experience teleoperation a robot arm.\footnote{Due to the COVID-19 pandemic and university restrictions, only those with pre-authorized access could participate. See the COVID-19 considerations document in the \supp{} for more details.} We conduct a within-subjects study, where each participant completed 2 tasks, chosen randomly, with each of the three methods. Users were given 2 trials to complete each tasks, and an allotted 3 minutes per control strategy to practice. Users were also given a sheet describing controller inputs and details for each control method, which we include in the \supp{}. For imitation learning and LILA controllers, which require language inputs, participants provided a natural language utterance which which a proctor entered into the model -- participants were allowed to verify the proctor entered their input accurately. This user-provided language utterance is used as the query in the nearest-neighbor retrieval described in \autoref{subsection:integrating-language}; the retrieval set consists of all training utterances collected via the crowdsourcing procedure above. In addition to tracking quantitative success rates (normalized, based on progress relative to each of the 4 defined subtasks), time taken per task, and controller logs, we ask users to fill out a qualitative survey evaluating each method at the end of each study. We present both quantitative and qualitative results below.

\paragraph{Quantitative Results.} \autoref{fig:quantitative} summarizes our objective results. We evaluate both full- and partial-task success rates for each task across all control methods, in addition to computing smoothness metrics directly on the logged user inputs and robot actions. Smoothness is a measure for intuitiveness when measured on user 2-DoF joystick inputs, ease of use when measured on the robot's end-effector pose, and implicit safety: a trajectory with high discontinuity in acceleration can lead to rapid, unpredictable changes in the environment. Smoothness is negatively correlated with \textit{jerk}, the time-derivative of acceleration. We compute jerk by taking the second-order derivative of velocity, and report average jerk across fixed windows. 

Our results show that LILA significantly ($p < 0.05$) outperforms both methods across all sub-tasks, and is also smoother to use both in control input space (2-DoF input) and end-effector space (6-DoF). However, the relative drop in performance of LILA for the final sub-task ``Complete Task'' shows the room for improvement in fine-grained control, such as pouring motions. Additionally of note is the poor performance of imitation learning. To explore this fully, we perform an ablation, and show that sample inefficiency is a likely cause -- especially since we are in a low-data regime. These results can be found in the \supp{}.\footnote{Additional experiment videos, and a results of the imitation learning ablation can be found here: \\ \url{https://sites.google.com/view/lila-corl21/home/il-ablation}}

\begin{figure*}
    \centering
    \includegraphics[width=\textwidth]{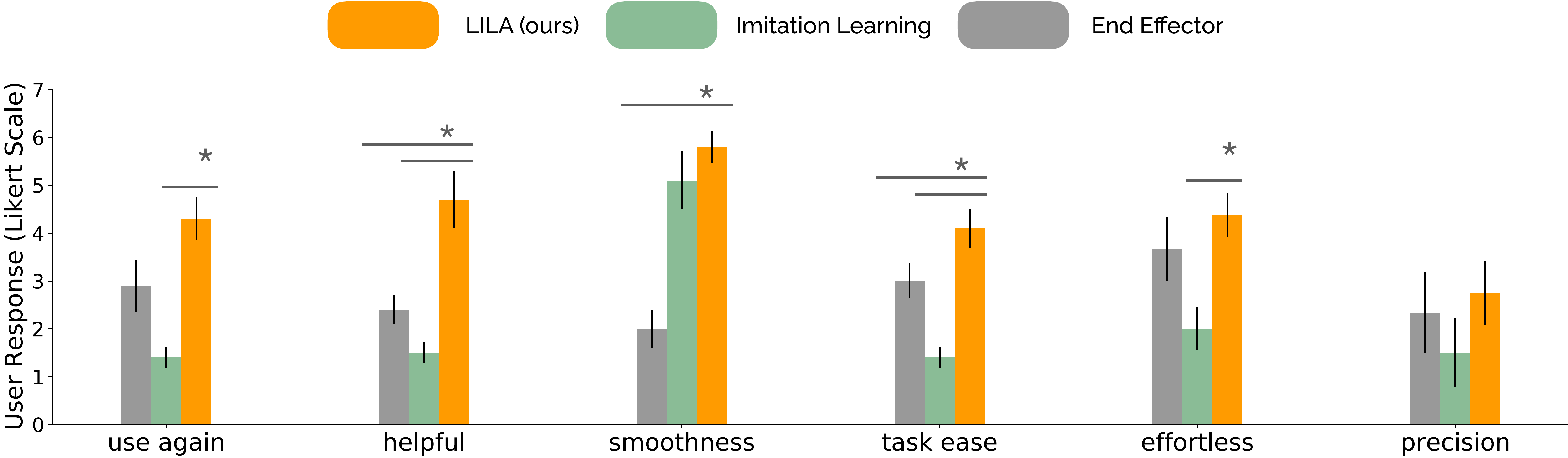}
    \vspace{0.03in}
    \caption{Qualitative Results. Using a 7-point Likert scale, we ask users to evaluate each of the 3 control methods for different properties.  With high significance ($p < 0.05$), we find that LILA outperforms both imitation learning and end-effector control baselines on several metrics, including degree of helpfulness provided and ease in completing tasks.}
    \label{fig:qualitative}
\end{figure*}

\begin{figure*}
    \centering
    \includegraphics[width=0.8\textwidth]{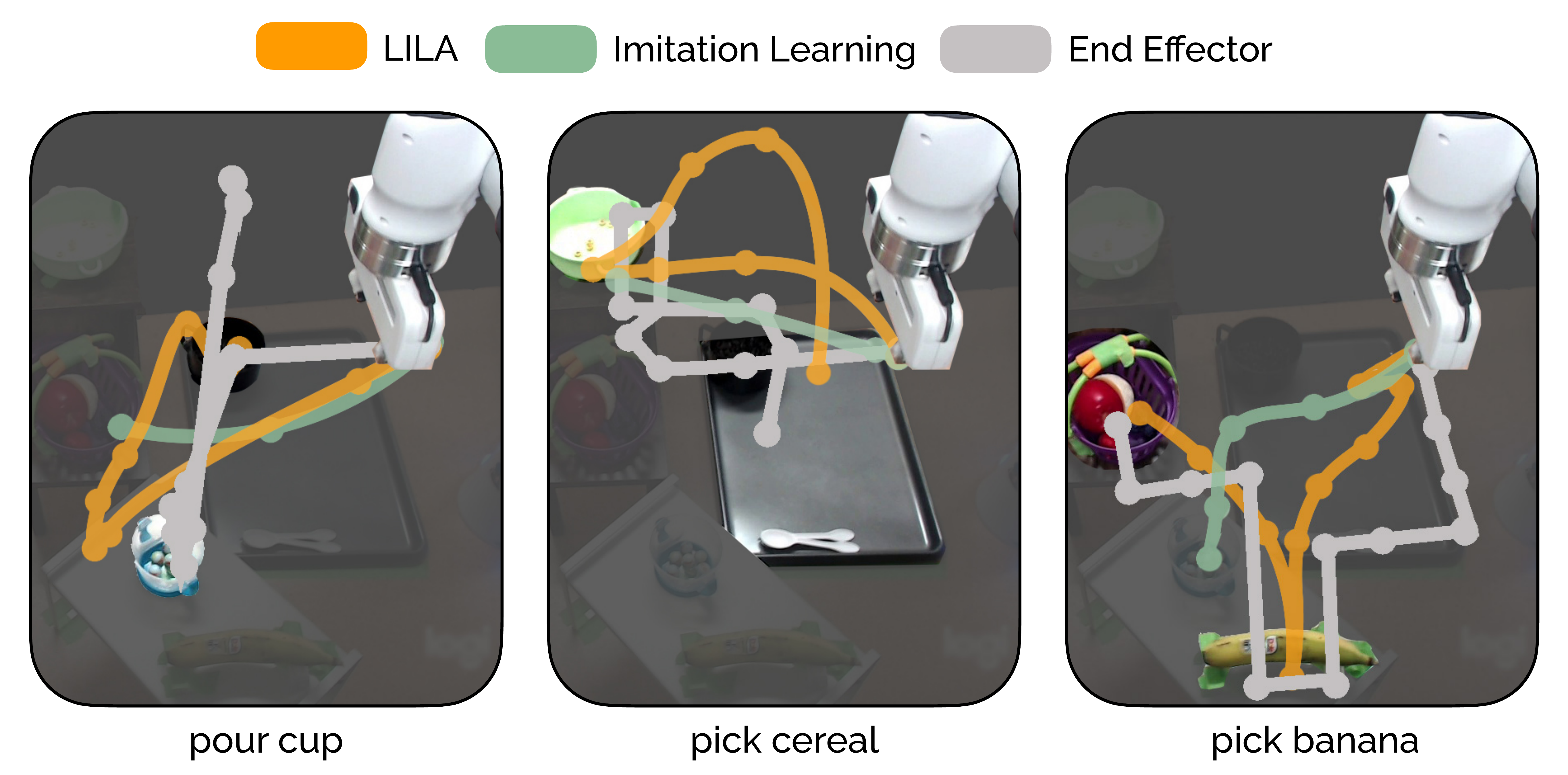}
    \vspace{0.03in}
    \caption{Trajectories from one user comparing three different control methods 3 out of our 5 tasks -- \textbf{\texttt{Pour Cup}}, \textbf{\texttt{Pick Cereal}}, and \textbf{\texttt{Pick Banana}}. LILA provides smooth actions that immediately approach the target object for a task, while end-effector control shows more rigid motions that result in users diverging from their intended paths. While imitation learning also enables smooth motions, it often fails shortly after reaching objects, hence the shortened trajectories.}
    \label{fig:user-traj}
\end{figure*}

\paragraph{Qualitative Results.} \autoref{fig:qualitative} summarizes our subjective results. We administered a 7-point Likert scale survey after users finished performing tasks with each method; this survey included questions around the perceived helpfulness of the model in completing the tasks (\textit{helpful}) and whether the participant would use the control method again (\textit{use again}). The results show that LILA outperforms both imitation learning and end-effector control across most qualitative metrics, with significant results ($p < 0.05$) marked with an $*$. We additionally visualize samples of the observed end-effector trajectories by individual users collected during our study for 3 of our 5 tasks in \autoref{fig:user-traj}. Across all tasks, LILA results in smoother end-effector trajectories than end-effector control, while imitation learning comes close to the target object but is unable to complete the entire trajectory for the task.  

\vspace*{-0.07in}
\section{Discussion}
\label{sec:discussion}
\vspace*{-0.03in}

\paragraph{Summary.} We present Language-Informed Latent Actions (LILA) a framework that marks the first step in combining the expressiveness and naturalness of language for \textit{specifying} and \textit{executing} on a human user's objective within the context of assistive teleoperation. Our user study results show that when compared to fully autonomous, imitation learning approaches, LILA is more sample-efficient and performant, training on half the number of task demonstrations, but obtaining significantly higher success rates. Compared with no-learning end-effector control methods, we again show LILA's effectiveness at obtaining high success rates, but also demonstrate its ability to produce intuitive low-dimensional control spaces from language input. Qualitatively, we find that users prefer LILA to alternative methods across the board, opening the door for additional work on language \& latent actions.

\paragraph{Limitations and Future Work.} Currently, LILA uses language as a mechanism for \textit{task disambiguation} -- in the current results, there is no mechanism for generalizing to completely unseen tasks or language specifications. We believe that the ability to disambiguate with language, and the integration of language within the latent actions framework is a strong research contribution, and hope that future work looks to dynamic states -- perhaps by leveraging visual latent actions \citep{karamcheti2021vla} -- and to adapting to new utterances and tasks dynamically \citep{karamcheti2020decomposition, wang2017naturalizing}. Furthermore, while users found LILA intuitive and natural, they found themselves wanting to further modulate the robot's behavior with language instructions \textit{during the course of execution}. Many users, upon seeing the robot make slight deviations from a desired path would instinctively provide spoken corrections -- \textit{``a little to the right''}, \textit{``no, grasp it by the handle!''} -- indicating a desire for \textit{multi-resolution} language control.

\paragraph{Shared Autonomy and LILA.} LILA fits within the shared autonomy paradigm, where the role of language has been underexplored. With LILA and shared autonomy approaches in general, humans retain \textit{agency} -- they are responsible for robot motion, and if the robot moves in a way that is not safe, or does not align with their objectives, they stop providing control and possibly reset, give a new instruction, or drop into a more complex control mode. Behavior is \textit{interpretable} -- the latent actions model, \textit{critically informed by language}, produces intuitive control spaces that humans can quickly grasp. Finally, language is \textit{natural} -- users specify their objectives as they would if speaking to another person, and the robot uses that language to shape their control space. These properties -- preserving agency, maintaining interpretability, and leveraging the expressive and natural features of language for specifying objectives -- are critical for widespread human-robot collaboration, and we hope this work presents a concrete step towards achieving that goal. 
	

\newpage
\acknowledgments{Toyota Research Institute (``TRI'') provided funds to assist the authors with their research but this article solely reflects the opinions and conclusions of its authors and not TRI or any other Toyota entity. This project was also supported by NSF Awards 2006388 and 2132847. Siddharth Karamcheti is grateful to be supported by the Open Philanthropy Project AI Fellowship. Megha Srivastava is supported by the NSF Graduate Research Fellowship Program under Grant No. DGE-1656518.

We would additionally like to thank the participants of our user study. We further extend a special thank you to Madeline Liao and Raj Palleti for helping with the visual intuition for our imitation learning analysis. Finally, we are grateful to our anonymous reviewers, Suneel Belkhale, Ajay Mandelkar, Kaylee Burns, and Ranjay Krishna for their feedback on earlier versions of this work.}


\bibliography{refdb}

\newpage
\appendix

We provide further details, experiments, and descriptions of the attached media, to reinforce the results and conclusions from the main body of our paper. For a more fluid viewing experience please look through our project website, where videos (and corresponding descriptions) are side-by-side: \url{https://sites.google.com/view/lila-corl21}. The top-level page contains videos from our actual user study, showing the various models in practice, while the sub-pages contain additional experiments exploring the poor performance of the imitation learning baseline in our work, as well as justifications for the omission of the no-language latent actions baseline.

We additionally provide a full version of our code repository at the following url: \url{https://github.com/siddk/lila}, with a detailed README spanning the entire LILA pipeline from demonstration recording to training models, to deploying them on a robot.

\section{COVID-19 Impact Statement}
\label{sec:covid-impact}
\vspace*{-0.03in}
Due to the ongoing COVID-19 Pandemic, the ability to run larger-scale user studies, and even have reliable lab access for preparing experiments, was limited. We are in a University setting, and as such, were susceptible to University restrictions.

All members of the authorship team had to go through a COVID-19 safety training, frequent testing, as well as an official approval process to be granted permission to work in the robot lab. For User Study participants, we were mostly limited to those with pre-existing access to the Engineering Building (spanning both robotics and non-robotics students), as well as a limited number of other University students granted approval to access other nearby buildings. This limited our ability to launch a larger scale user study, with more than 10 participants from a more diverse population.

That being said, we strongly believe that our existing participant statistics -- 10 students (5 female/5 male, age range $23.2 \pm 1.87$ -- reflect a broader user pool. Coupled with the statistical significance of the results we have already collected, we feel that the User Study results remain compelling, and our conclusions hold. That being said, we would like to run additional studies once COVID-19 restrictions relax in our area.

\section{Related Work Discussion}
\label{sec:supplemental-related-work}
\vspace*{-0.03in}
The main body of our paper contains a cursory discussion of different approaches for language-informed robotics, spanning a multitude of full-autonomy solutions. While this discussion helps provide contrast for our \textit{shared-autonomy \& language} based approach, LILA, it does not do the existing work in the field justice, nor does it discuss the data, implementation, and feature-engineering considerations that are coupled with these different approaches. 

First, we discuss work that leverages structured logical forms as an intermediate representation for mapping language. The benefits of these forms are that they induce \textit{logical forms} -- functional, often programmatic -- representations of meaning, that can then be executed on a robot, either by directly formulating a plan (requiring a model of the environment), or learning a lightweight policy from data that can fulfill these logical forms. A perceived benefit of these types of approaches is their sample efficiency -- by leveraging a highly structured logical form and possibly hand-engineered features, one can learn the language to logical form mapping with 10s of examples. An example of this work is \citet{macglashan2015grounding} that learns reward functions given hand-engineered linguistic features; however, these reward functions are fed to a planner (requiring full knowledge of world dynamics -- a huge assumption) to generate robot behavior. Follow-up work by \citet{arumugam2017accurately} relax the hand-engineered language feature assumption by using more recent neural approaches, but still hinge on using planners. While these approaches are sample-efficient, many of the assumptions around planning and full dynamics are strong, and limit the potential of scaling this work (and for manipulation, are not necessarily straightforward!). 

Other work that relaxes the planning/dynamics assumption is \citet{duvallet2013imitation}; this work uses hand-engineered features on top of a popular logical form -- Spatial Description Clauses (SDCs) -- to learn policies for robot navigation. While seemingly as sample efficient as the prior approaches without the downsides, there is still a high cost for \textit{policy learning}. Though the approach only required 10s of examples to learn to map language to the appropriate SDC, learning an effective policy (note this is just discrete node navigation -- not continuous manipulation) required running DAgger \citep{ross2011reduction} for 25 iterations on top of their existing data, collecting an order of magnitude more demonstration data (or demonstration edits/corrections as in DAgger) than originally given -- 100s - 1000s of demonstrations.

On the other end of the spectrum are more recent approaches in end-to-end robot learning for more complicated tasks spanning navigation \citep{anderson2018vision, ku2020rxr} and manipulation \citep{shridhar2020alfred}. This work is done in simulation, where one has the ability to do virtually infinite policy rollouts given a language instruction paired with a reward function, to learn robust policies via reinforcement learning. Other work in this paradigm that uses imitation learning, but with real-world deployments include works for quadcopter flight \citep{blukis2018following} and some limited manipulation \citep{stepputtis2020lcil}. These works still require 1000s of demonstrations, but use smart tricks for data augmentation and synthetic generation to learn robust policies.

LILA hopes to fill a void between these two classes of approaches; retaining the sample-efficiency of the earlier approaches, without the need for hand-engineered features, strong assumptions about known dynamics, or limited generalization potential. The experiments in the main body show that LILA is \textit{extremely} sample efficient; however, the scope of this work is mostly using language as a means for \textit{disambiguation}. It is our ardent hope that future work in language \& latent actions (and shared autonomy more generally) turns to more dynamic settings, richer language, and hard forms of generalization; this work is just the first step.

Finally, we want to help fill out the story of language-informed robotics with work that does not necessarily involve \textit{execution} (learning a policy or control space for robots to follow language instructions), but that can help language-based methods generalize better, from less data, or that leverage other modalities to help with specification. For example, \citet{matuszek2014unscripted} learn to map \textit{unscripted} interactions consisting of language and gestures to object localizations from relatively few interactions; such methods are crucial for scaling up LILA to multiple objects, referring expressions, and compositional language instructions. 

Other work looks at other modalities like speech and gesture to learn logical forms that allow for efficient generalization \citep{kollar2013towards}. Other work combines several modalities on top of language like gaze, gesture, and intonation to further help lift representations of human intent from language \citep{kennington2013situated, whitney2016multimodal}. All this work, though not directly related to LILA in that they do not help learn meaningful control spaces, do present possible avenues through which we might scale this approach to new contexts, language instructions, and behaviors.

\begin{figure*}
    \centering
    \includegraphics[width=\textwidth]{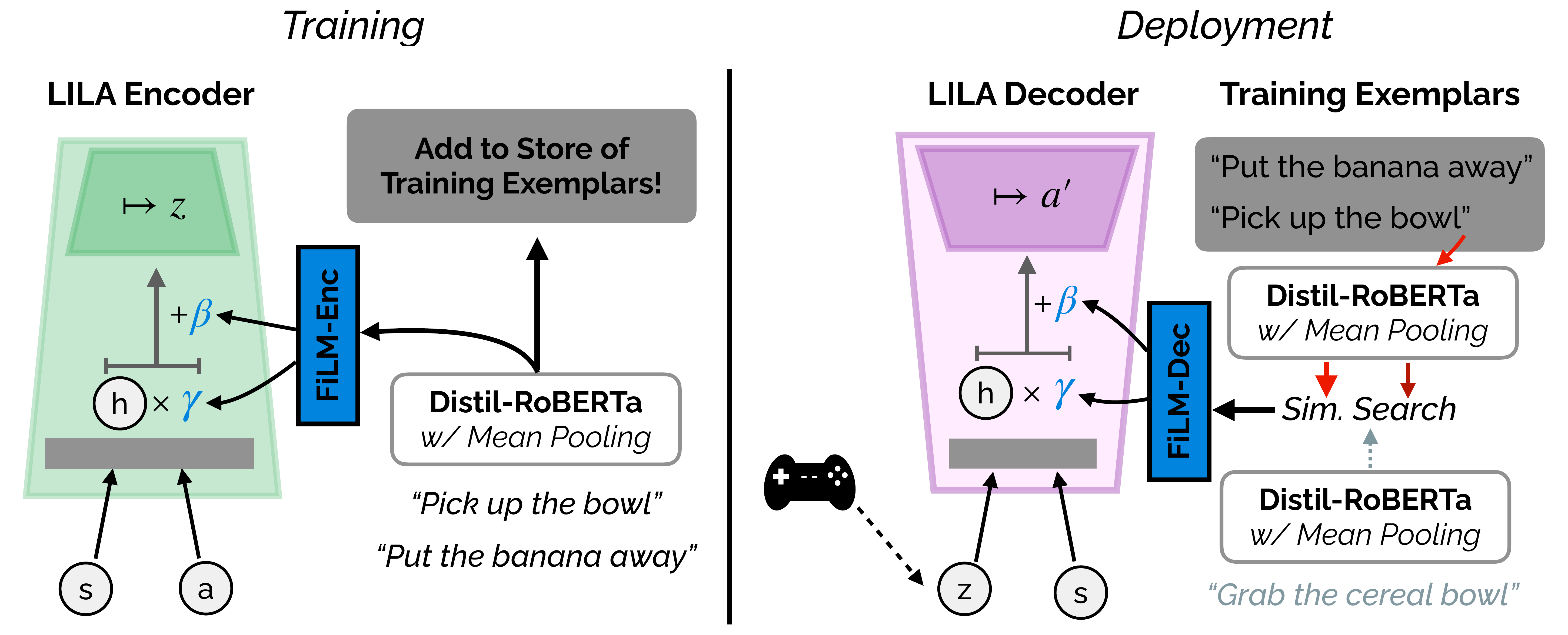}
    \vspace{0.03in}
    \caption{An overview of our language-informed latent actions (LILA) models with the FiLM modules highlighted. For additional clarity, we provide partial views of both the Encoder (during training) and Decoder (during deployment).}
    \label{fig:supplement-arch}
\end{figure*}

\section{Model Architectures \& Training}
\label{sec:architecture-training}
\vspace*{-0.03in}
Following from the main body of the paper, we provide additional details on the model architectures, and additional data processing/augmentation we use in this work. We start with a more thorough description of the feature-wise linear modulation (FiLM \citep{perez2018film}) mechanism we use to integrate language into the latent actions pipeline.

\paragraph{FiLM Architecture for LILA.} Prior work in latent actions \citep{losey2020latent,karamcheti2021vla} implement the latent action models as simple feed-forward multi-layer perceptrons (MLPs) with \texttt{tanh} activations as the non-linearity between layers. The Encoder and Decoder are symmetrical, with the Encoder encoding a combination of $(s, a)$ pairs down to the latent space $z$ (via an intermediate layer or two), and the decoder decoding from $(s, z)$ back up to the 7-DoF action $a$. The intermediate layers in both the Encoder and Decoder have the same dimensionality of $30$ -- we refer the reader to consult the attached code for more detail.

With LILA, the two differences are that 1) we use the \texttt{GELU} activation instead of the \texttt{tanh} due to its better stability, and 2) we incorporate language into this existing pipeline. Recall that we use a version of the pretrained Distil-RoBERTA language model \citep{reimers2019sentence, reimers2020multisbert} to generate embeddings of each user utterance. These embeddings have dimensionality of $768$, which far exceeds the dimensionality of the 7-DoF $(s, a)$ of the typical pipeline. Initial experiments attempting to naively concatenate the language embedding with the 7-DoF states and actions (or states and $z$ values, in the case of the decoder) were not fruitful, as we were unable to learn (loss failed to decrease when training).

Instead, we turned to FiLM. The core principle with FiLM is that it's a fusion mechanism that does not increase the intrinsic parameter count of the core neural network (e.g., the original Latent Action MLP described above). FiLM has found great success in many multi-modal tasks for this reason, integrating with pre-existing pipelines in image classification to enable visual-question answering \citep{perez2018film}, as well as integrating into existing pipelines for instruction following in reinforcement learning \citep{chevalierboisvert2019babyai, bahdanau2019reward}. FiLM works as follows: given an intermediate representation from the Encoder or Decoder $h$ with dimensionality $d$ (say after the first layer of the corresponding MLPs), and a language embedding $e$, FiLM works in the following fashion:
\begin{align*}
    \text{FiLM-Gen}_\theta(e) &= \gamma_e, \beta_e \\
    h' &= \gamma_e \odot h + \beta_e
\end{align*}
where $h'$ is the representation fed to later layers in the Encoder/Decoder MLP, and $\odot$ denotes the Hadamard product (component-wise multiplication). Simply put, $\text{Film-Gen}_\theta$ is a module that learns to \textit{shape} the representations learned by the core Latent Actions MLP, injecting language information through this affine transformation defined by $\gamma_e, \beta_e$. We implement $\text{Film-Gen}_\theta$ as a separate two-layer MLP that also uses the \texttt{GELU} activation. \autoref{fig:supplement-arch} breaks this down visually, showing how the FiLM modules are added for both the Encoder and Decoder.

\paragraph{Imitation Learning Architecture.} For Imitation Learning, we do not have this Encoder-Decoder structure, with two separate FiLM modules. Instead, Imitation Learning is implemented as a single MLP (of same parameter count/number of layers as the Encoder + Decoder in LILA) that conditions on the state $s$; we add a single FiLM module after the first-layer of this MLP.

\paragraph{Data Augmentation.} There are two key aspects to our data augmentation procedure: 1) enforcing latent action consistency, and 2) adding robustness to noise.

A key desire in latent action models is consistency in nearby states -- executing the same latent action $z$ in nearby states should be roughly similar. More formally, $d_T(\mathcal{T}(s_1, \phi(z, l, s_1)), d_T(\mathcal{T}(s_2, \phi(z, l, s_2)))) < \epsilon$ for $\| s_1 - s_2 \| < \delta$, for some $\epsilon, \delta > 0$, where $d_T$ is some distance metric (e.g., Euclidean distance between states). We enforce this with a sliding window approach; given a sequence of states within a fixed window size, we train the decoder to predict the same actions given the $(z, l, s)$ triples for all states within the window.

The second, and most important augmentation we do is adding robustness to noise. Though we collect a dataset of $(s, l, a)$ triples, we use a unique property of our control space to obtain better robustness: because our states are the actual joint states (let's call this $q$), and actions $a$ are just joint velocities ($\dot{q}$), we can ``re-compute'' actions between two sequential states $(s_1, s_2)$ by just taking the finite difference $a' = s_2 - s_1$. This allows us to do the following: add noise to each initial state $s_i$ subject to $\epsilon = \mathcal{N}(0, \sigma)$ (we use $\sigma = 0.01$), then compute the ``corresponding action'' by taking $a' = s_{i + 1} - (s_{i} + \epsilon)$. This can be viewed as a simulated version of the DART paradigm \citep{laskey2017dart} for noise-robust imitation learning.

We train LILA models with both the above augmentations. While the former augmentation mode is not directly applicable to Imitation Learning approaches, the second noise augmentation mode is -- indeed, we find we have to triple the amount of such augmentations, in order to get even slightly meaningful behavior from Imitation Learning models.

\section{Demonstration Collection}
\label{sec:demo-collection}
\vspace*{-0.03in}
\begin{figure*}
    \centering
    \includegraphics[width=0.8\textwidth]{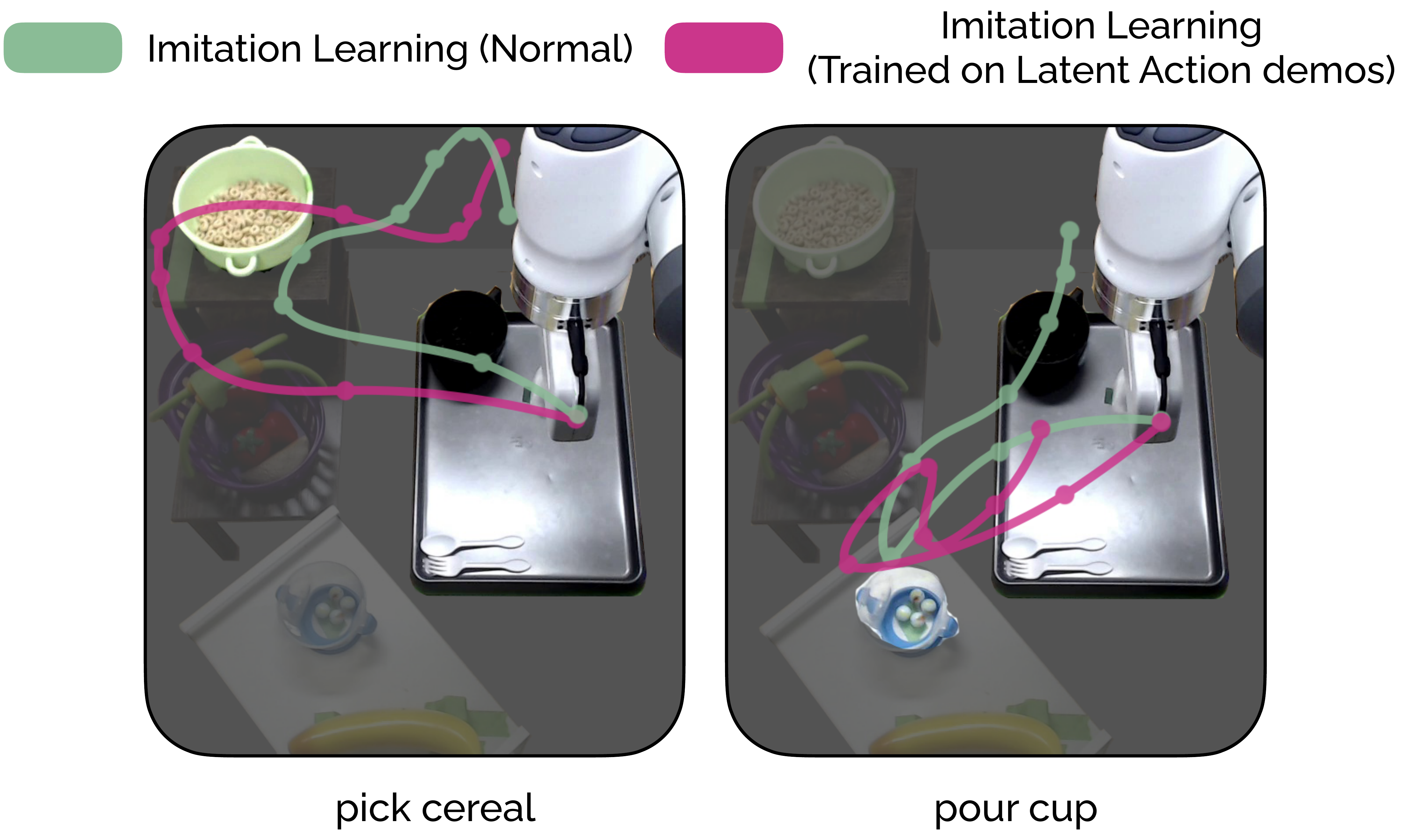}
    \vspace{0.03in}
    \caption{Visualized Trajectories for an Imitation Learning model trained on the ``pure'' data (that is reported in the paper), and an Imitation Learning model trained on the ``LILA-style'' demonstrations with the ``sweeping'' motions. On the left are trajectories for the instruction \textit{``grab the cereal bowl''} and on the right, \textit{``pour the blue cup into the black coffee mug''} -- these two examples are from our training language set. We observe that Imitation Learning trained on the ``LILA-style'' demonstrations performs slightly worse by not following the complete ideal task trajectory, although neither approach is able to successfully complete the task.}
    \label{fig:il-vs-la-demos}
\end{figure*}

Both Latent Action models and Imitation Learning models require a dataset of language paired with corresponding demonstrations to perform learning. As mentioned in the main body of the paper, we collect these demonstrations \textit{kinesthetically}, manually moving the robot arm to complete specific tasks, recording the joint states and actions along the way. Critically, for imitation learning, these demonstrations consist solely of what we call the ``forward'', or ``pure'' demonstration of a task: starting at the home position, perform each of the individual subtasks smoothly and continuously, until the task has been satisfied. For a task like ``put the banana in the fruit basket'' this corresponds to 1) smoothly reaching for the banana and grasping it, 2) lifting the banana up and moving over to the basket, and 3) inserting the banana into the basket and releasing the gripper.

However, when collecting demonstrations for latent action models, we find that we can learn \textit{better, more reliable} models by changing up the demonstration process a little, incorporating discrete segments of the demonstration where we \textit{back off and then repeat} a motion. Concretely, for a task like ``put the banana in the fruit basket'' this corresponds to 1) smoothly reaching for the banana, pulling back to the home position, \textit{and then} reaching for the banana again and grasping it, 2) lifting the banana up, lowering it, then taking it to the basket, and 3) finally dropping the banana in the basket. These ``sweeping'' back-and-forth motions intuitively help the latent actions model induce control spaces that give users \textit{reversible} control -- the ability to move the robot back, rather than just press forward; this is critical to usability and recoverability.

\paragraph{On the Fairness of Comparing Imitation Learning and LILA.} Because LILA and Imitation Learning are trained with different demonstrations, there is a plausible fear that our comparison between LILA and Imitation Learning is unfair -- specifically, because the latent actions demonstration incorporates extra motion, LILA technically may be seeing more $(s, a)$ pairs per demonstration compared to Imitation Learning, and can therefore learn better.

We mitigate this in two ways; first, as a side effect of doubling the number of collected demonstrations for Imitation Learning -- recall that, in the main paper, to get IL to show semantically meaningful behavior, we needed to collect 30 demonstrations per task instead of the 15 per task LILA was given -- we found that Imitation Learning sees $40\%$ \textit{more} $(s, a)$ pairs than LILA (even more if you account for the fact that we tripled the data augmentation for Imitation Learning!). Second, \autoref{sec:additional-experiments-il} presents a more concrete experiment where we train the Imitation Learning model \textit{on the LILA demonstrations}, and show that the resulting model performs worse than when trained on the standard IL dataset.

\begin{figure*}
    \centering
    \includegraphics[width=0.7\textwidth]{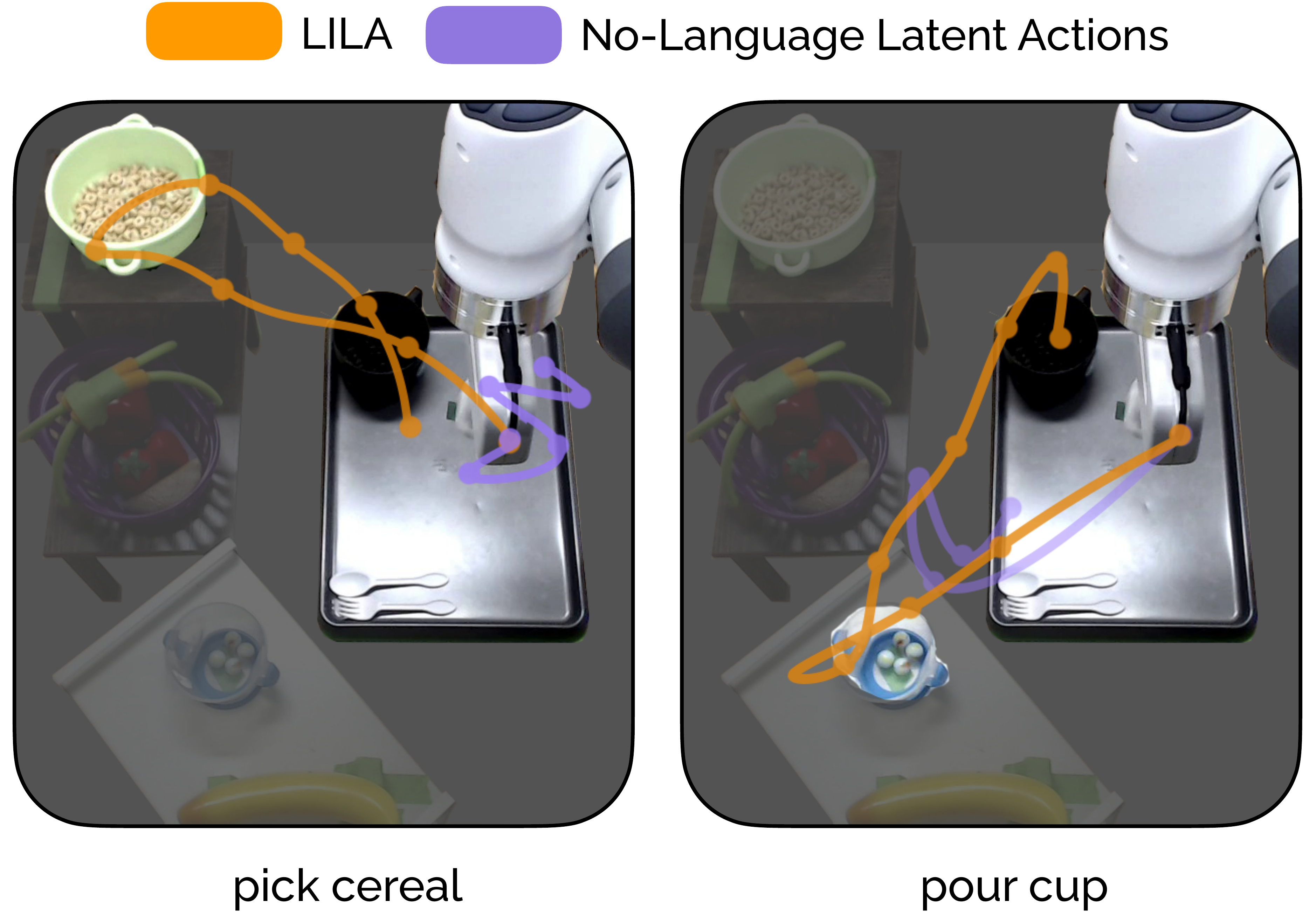}
    \vspace{0.03in}
    \caption{Visualized Trajectories for a No-Language Latent Actions model vs. LILA (with language) trained on the same set of demonstrations, operated by an expert user. With the No-Language Latent Actions model, the user tries their best to complete the task with the provided controls.  On the left are trajectories for the instruction \textit{``grab the cereal bowl''} and on the right, \textit{``pour the blue cup into the black coffee mug''} -- these two examples are from our training language set. We observe that the No-Language Latent Actions is unhelpful for completing the task, and unable to even completely reach the target object for either task, demonstrating the importance of incorporating language to help condition the learned latent actions.}
    \label{fig:no-lang-vs-lila}
\end{figure*}

\section{Additional Experiments: Imitation Learning Baseline}
\label{sec:additional-experiments-il}
\vspace*{-0.03in}
A critical question from our user study has to do with the poor performance of the imitation learning baseline relative to both LILA and End-Effector control. This section explores additional ablation experiments as well as a technical argument for why imitation learning performs poorly -- namely due to sample inefficiency, exacerbated by our real-robot setting.

We also address the point raised in \autoref{sec:demo-collection} -- how Imitation Learning performs when trained with the ``latent actions'' style demonstrations (``sweeping'' motions) rather than the ``pure,'' straight-through demonstrations.

\paragraph{Ablation Experiments.} The following URL contains several sets of ablation experiments, with corresponding text annotations: \url{https://sites.google.com/view/lila-corl21/home/il-ablation}. Many of these experiments show qualitative behavior, and are best viewed via the linked URL; however, we summarize the main findings here.

First, we put LILA and Imitation Learning on an equal footing, picking 3 of the 5 original tasks we
used in the user study -- namely, \textbf{\texttt{Pick Cereal}},  \textbf{\texttt{Pick Fruit Basket}}, \textbf{\texttt{Pour Cup}}. We collect 10 demonstrations for each task, and trained a base model with the noise-based data augmentation (add noise once for each state), for both LILA and Imitation Learning. We show that LILA is able to fully succeed at all three of these tasks with \textit{only 10 demonstrations}, whereas Imitation Learning completely fails. However, we note that even at 10 demonstrations, Imitation Learning is starting to show semantically meaningful behavior -- for the \textbf{\texttt{Pick Cereal}} the end-effector clearly moves toward the cereal bowl (though not close enough to grasp), and then down to the tray (though not close enough to execute a successful drop).

We then experiment with the impact of data augmentation, first looking at 3x the amount of augmentation (noising each state 3 times in the dataset following the procedure detailed above), and then 5x. We show that Imitation Learning performance is slightly better than the reference with 3x data augmentation, but that 5x doesn't additionally help.

We then vary the number of demonstrations from 10 to 20, then 30 demonstrations per task, with 3x data augmentation. Critically, we show that Imitation Learning \textit{improves as we add more demonstrations, even though it still isn't able to solve the tasks.} As an interesting data point, at 20 demonstrations, the Imitation Learning agent is able to execute a successful grasp of the cup (see the video on the webpage), but cannot finish the task. 

Unfortunately, after collecting 90 demonstrations (30 for each task), we felt we'd need close to 50-100 demonstrations per task to actually get imitation learning to solve these tasks -- almost \textit{two orders of magnitude} more data than LILA needed. This would have been prohibitive to collect so we stopped, and instead decided to analyze the possible cause of this extreme sample inefficiency. Again, videos and annotations depicting these ablations visually, in an easy-to-follow format can be found here: \url{https://sites.google.com/view/lila-corl21/home/il-ablation}.

\begin{figure*}
    \centering
    \includegraphics[width=\textwidth]{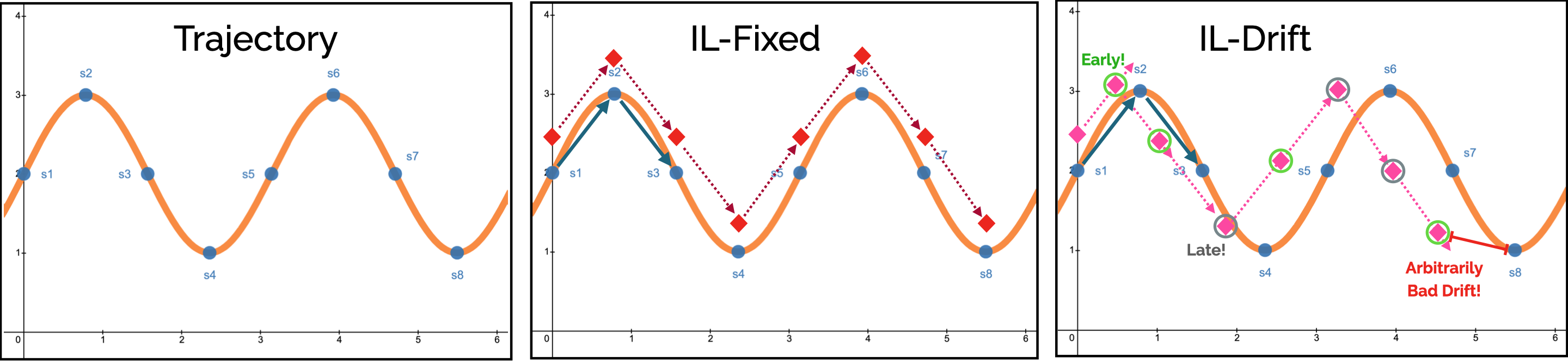}
    \vspace{0.03in}
    \caption{We consider a sinusoidal trajectory of a simplified end-effector through 2D space, where a robot's motion is continuous. If we can sample states when collecting this ``demonstration" at a fixed interval, then the state-error of an imitation learning agent trained with behavior cloning grows minimally over time. However, the fixed interval assumption may not be true when collecting data on a real robot, where any noise can lead to compounding errors resulting in arbitrarily bad drift. This example supports our observation of poor IL performance on the 7-DoF Franka Emika Panda robot in our user study.}
    \label{fig:il-results}
\end{figure*}

\paragraph{Why is Imitation Learning Sample Inefficient?} The above experiments point to an extreme sample inefficiency in Imitation Learning. Imitation Learning is generally subject to problems of cascading errors and the general noise problems of real robotics (imprecise resets, inherent noise in robot joints). However, we will now present an argument that illustrates why Imitation Learning and LILA may be even worse than expected in our real-robot setting, due to 
specific implementation details in our publish/subscribe based methodology. Notably, there is imperfect communication between the top-level Python process (housing the learned models) and the low-level C++ robot controller that exacerbates the cascading errors imitation learning has to deal with. A graphical walkthrough of this argument can be found at the bottom of the webpage for this section here: \url{https://sites.google.com/view/lila-corl21/home/il-ablation}.

Consider the sinusoidal trajectory of a simplified end-effector through 2D space shown in Figure \ref{fig:il-results}. The robot's motion is continuous, but we can sample states when collecting this "demonstration" at fixed intervals.
Notably, this is an assumption present implicitly in most simulators (fixed frame rate, or fixed control iterations/sec) – this is also usually true when operating with discrete action spaces (move forward/right/left).
However, this fixed interval assumption may not be true when collecting data on a real robot, depending on the implementation. More on this in a bit, but for now, let's assume our demonstration data consists of evenly spaced (state, action) pairs with this fixed interval between samples.

Consider training an imitation learning agent via behavioral cloning, where the policy is parameterized as a neural network. It's not clear what a NN will do given a state outside it's training distribution (could be arbitrarily bad) , but to simplify, \textit{let's assume given a new state, this policy will predict an action based on retrieving the "nearest-neighbor" from it's training demo.} Assume there's some noise in the reset (this is representative – there's always *some* noise in the initial joint states - this is represented in simulators like Mujoco and PyBullet). If you roll out the imitation learning policy, you get behavior like that shown in the middle of Figure \ref{fig:il-results} – \textit{critically, assuming the same constant sampling rate, the state-error grows minimally over time.} However, for continuous state and continuous action robotics grounded in a real-world robot, this assumption does not exist, which leads to the following point.

With our implementation on a 7-DoF Franka Emika Panda robot, we noticed that despite our best efforts, \textit{we are not able to ensure states/actions are dispatched at a constant sampling rate}. The result (based on our straw man nearest-neighbors NN argument from above – though note the real world behavior, especially in higher dimensions will be much much worse!) is shown in Figure \ref{fig:il-results} on the right.  With slippage in the read/publish times of states and actions, we can read states too early or too late, execute a "bad" action, and \textit{cascade to arbitrarily bad final states over the course of execution} (even completely breaking in the middle of execution). 

\begin{figure*}
    \centering
    \includegraphics[width=\textwidth,height=125pt]{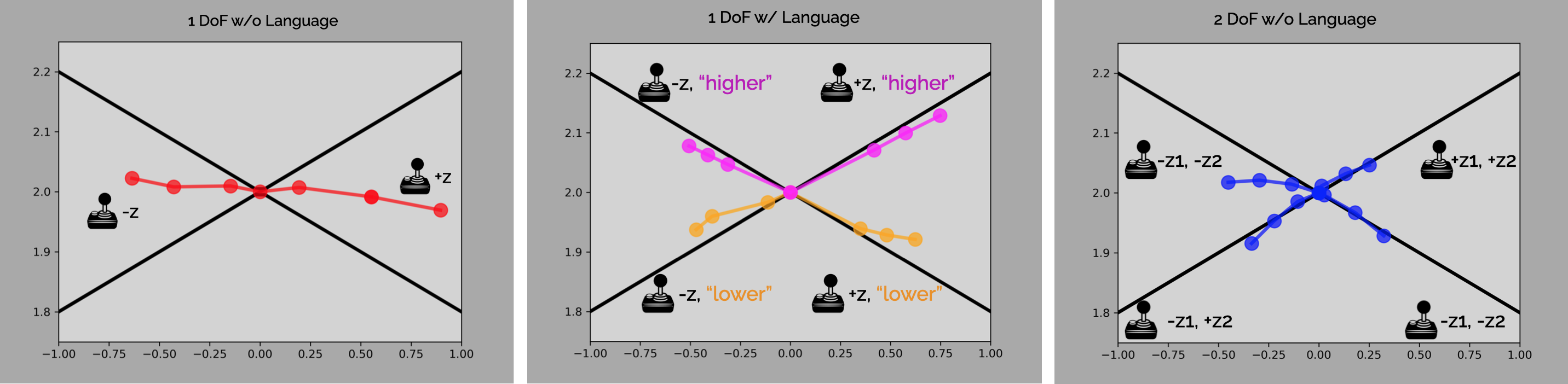}
    \vspace{0.03in}
    \caption{We train 3 different latent action models for the cross disambiguation task from Figure 2 of the main paper, which showed a simple ''cross" example, where starting from the mid-point, the goal is to be able to navigate towards all 4 possible directions. As the standard LA framework only takes in the controller inputs, it is impossible for a user to succeed with only 1 degree-of-freedom (DoF). To solve this task, we need an additional axis to condition on (at least 2-DoF). Our results show that, as expected, without any additional input such as language, a 1-DoF controller (left) is incapable of task disambiguation, with a clear 0\% task success rate for our simple cross setting.}
    \label{fig:nolang-results}
\end{figure*}

\paragraph{Imitation Learning with Latent Action Demonstrations.} Finally, as raised in \autoref{sec:demo-collection}, there is a question about the fairness of training on the ``LILA-style'' demonstrations with the ``sweeping'' motion vs. the more pure, forward-only imitation learning demonstrations. To address this, we train two Imitation Learning models (identical architecture, augmentation), with one model trained on the original ``pure'' data (the model from the main paper, used in the user studies) and a model trained on the ``LILA-style'' demonstrations with the sweeping behavior.

\autoref{fig:il-vs-la-demos} shows visualizations of the trajectories for the models rolled out on two instructions from the training set. We see that the Imitation Learning model trained on the LILA demonstrations is worse than the ``standard'' Imitation Learning model. Whereas the pure model is able to at least closely reach the target objects, the other model attempts to reach for an object, but wastes time moving around the object rather than focusing on the task trajectory -- intuitively this makes sense, as the ``sweeping'' motions present in the LILA data are confusing for imitation learning; given two opposing actions in the same state, what should it do? 

These experiments, coupled with the experiments in the main paper provide ample evidence that the comparison between LILA and Imitation Learning is not only fair, but highlights the sample efficiency of LILA as well.

\section{Additional Experiments: No-Language Ablation}
\label{sec:additional-experiments-nolang}
\vspace*{-0.03in}
Another possible question is how latent actions performs \textit{without language} -- in other words, is LILA necessary, or are prior latent actions models expressive enough to solve the tasks?

We answer this in two ways. First, we train a latent actions model, completely ablating the language encoding pipeline (keeping architecture the same, but removing the FiLM components described in \autoref{sec:architecture-training}). The corresponding latent action decoder only takes in the latent action $z$ and state $s$ as an input to predict high-DoF actions $a$.   \autoref{fig:no-lang-vs-lila} shows visualizations of trajectories for this No-Language model as well as LILA operated by an expert user, trying to accomplish two specific tasks. While LILA performs as expected, the No-Language model is unable to make progress at all. As soon as control begins and the user moves the robot into a state close to any object, the control space loses meaning, and the user is unable to recover, instead generating random behavior. Again this makes sense; without language to condition on, the No-Language model has no idea what task to perform. It lacks the ability to \textit{disambiguate} tasks, and as such, cannot make progress. We present additional videos and experiments to the above here: \url{https://sites.google.com/view/lila-corl21/home/no-lang-baseline}.

Second, we train 3 different latent action models for the cross disambiguation task seen in Figure 2 of the main paper:  A 1 DoF without language baseline, LILA (our proposed approach) with 1 DoF, and a 2 DoF without language baseline.  Each model is trained on a dataset of 100 demonstrations collected across the 4 tasks. We then visualize movement trajectories (see \autoref{fig:nolang-results}) by controlling the latent action ($z$ for 1 DoF, $z_1$ and $z_2$ for 2 DoF). As expected, without any additional input such as language, a 1-DoF controller is incapable of task disambiguation, with a clear 0\% task success rate for our simple cross setting.  With both our 1-DoF controller w/ language input and a 2 DoF controller, task disambiguation is possible – highlighting the necessity of additional information of any modality. However, note that in large multi-task environments with dozens of tasks, re-designing a controller can be difficult – language is a much more flexible and natural way to add this information.

Together, these results motivate why no-language latent action baselines are incapable of being useful for multi-task environments, as they are limited by total degrees of freedom of the controller. Because the goal of our user study is to compare methods that could be useful for successful task completion, If included, such a baseline would be uninformative as it would be impossible for any user to achieve above a  0\% task completion rate.

\section{Crowdsourcing \& User Study}
\label{sec:crowd-studies}
\begin{figure*}
    \centering
    \includegraphics[width=\textwidth]{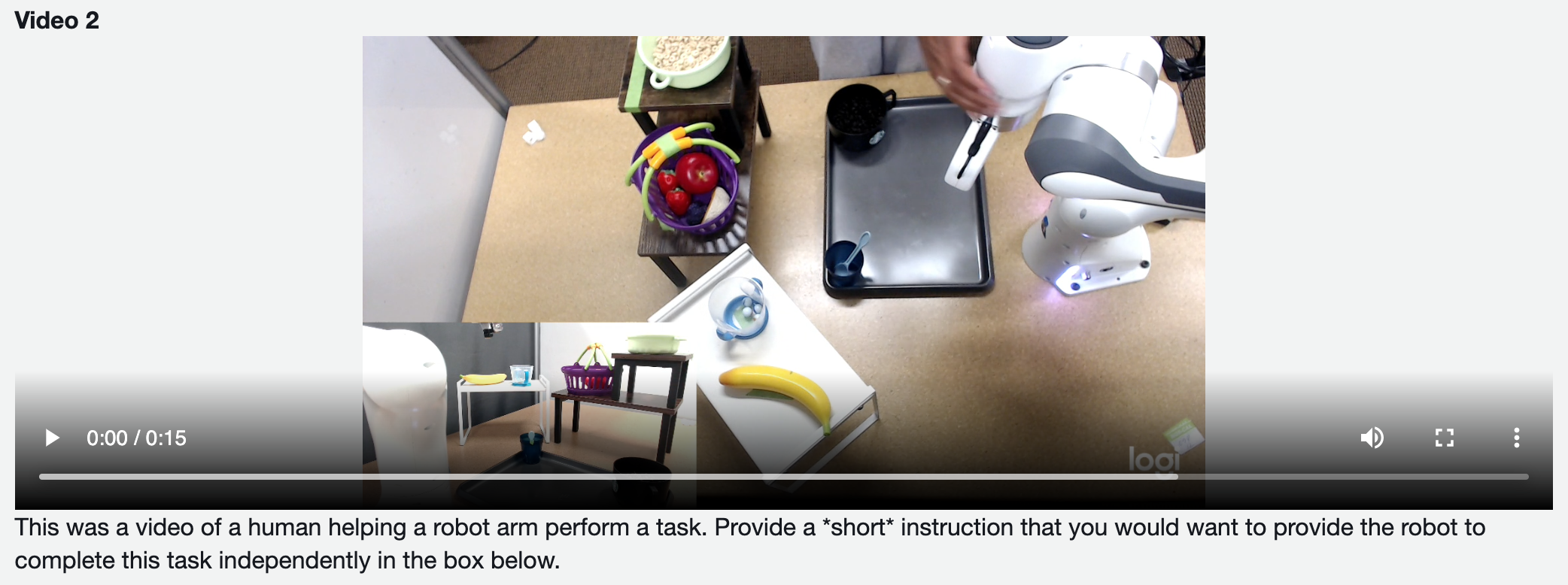}
    \vspace{0.03in}
    \caption{Interface shown to crowdworkers for collecting language utterances.}
    \label{fig:mturk}
\end{figure*}

\vspace*{-0.03in}
As described in the paper, to train LILA with language utterances, we hire crowdworkers on Amazon Mechanical Turk to provide language utterances give video demonstrations of a human assisting the robot arms. We paid crowdworkers $1.20$ dollars to provide short utterances for \textit{seven} videos. Notably, crowdworkers were not given any information about the possible tasks, or names of the objects in the scene. An image of the interface provided to crowdworkers is included in \autoref{fig:mturk}.

\begin{algorithm}[t]
	\caption{Filtering Language Utterances} 
	\begin{algorithmic}[1]
		\For {{\tt task} $=1,2,\ldots, T$} 
		    \State Initialize list $E_{\text{task}}$. 
			\For {{\tt user} $=1,2,\ldots,N$} 
				\State Append embedding of utterance {\tt embed(user, task)} to $E_{\text{task}}$
			\EndFor
		\EndFor
	    \For {{\tt user} $=1,2,\ldots,N$}
	        \State Initialize list $D_{\text{user}}$
	    \For {{\tt task} $=1,2,\ldots, T$} 
	        \State Append {\tt cosine distance} $d$ between {\tt embed(user, task)} and  {\tt avg}$(E_{\text{task}})$ to $D_{\text{user}}$
	    \EndFor
	    \EndFor
	    \State Filter out all utterances from users with $K$ highest {\tt avg}$(D_{\text{user}})$  
	\end{algorithmic} 
\end{algorithm}

\paragraph{Filtering Crowdsourced Language Annotations.} As described in the main paper, one issue with crowdsourcing language utterances from Amazon Mechanical Turk is the presence of ``spam'', noise, or extremely out-of-domain annotations. We accept and pay all crowdworkers for their responses, but adopt the following filtering algorithm before training on the collected utterances:

We set the threshold K to use utterances from 15 different crowdworkers for each of the 5 tasks. Example utterances from crowdworkers who we filtered out include \textit{``move your arm outwards towards the left about 10 inches, lower your arm until you can't anymore then close your claw.''} and \textit{``the human the robot to take bowl''}, highlighting the challenges of eliciting high quality language utterances from demonstrations, and the necessity of such filtering methods.

All 15 utterances for each of the 5 tasks are listed in the file \texttt{`full-annotations.txt'} (under \texttt{attachments/} in the code repository), constituting the entirety of our training data. The file \texttt{`filtered.txt'} contains the utterances that were filtered out by our procedure. 

\paragraph{User Study Instructions.} As described in the main paper, we conducted a user study with 10 participants. The attached PDF file \texttt{`user-study.pdf'} shows the example instructions provided to each participant in our study.

\end{document}